%% file: main.tex
\documentclass[twocolumn]{svjour3}
\pdfoutput=1
\usepackage{mathptmx}

\usepackage{xspace}
\usepackage{graphicx}
\usepackage{amsmath}
\usepackage{amssymb}
\usepackage{hyperref}
\usepackage{color}

\def\figdir{tinyfigs}

\def\Vec#1{\ensuremath{\mathbf{#1}}\xspace}
\def\Fig#1{Figure \ref{fig:#1}\xspace}

\def\centre{\ensuremath{\Vec{c}}\xspace}

\def\Trn{^{\mathsf T}}									%Matrix transpose
\def\Deg{$^{\circ}$\xspace}
\def\rtwo{\ensuremath{\mathbb{R}^2}\xspace}

\def\markchange{}

\begin{document}
%\def\makeheadbox{}
%\date{}
%\title{Plumb-line camera distortion self-calibration using using minimal Hough entropy.}
\title{Camera distortion self-calibration using the plumb-line constraint and minimal Hough entropy.}

\author{Edward Rosten \and Rohan Loveland}

\institute{Edward Rosten\and Rohan Loveland\at
%ISR-2 Space and Remote Sensing\\
Los Alamos National Laboratory\\
Los Alamos, New Mexico, USA\\
\email{edrosten@lanl.gov $\cdot$ loveland@lanl.gov}
\and
{\markchange
Edward Rostem\at
Department of Engineering\\
University of Cambridge, Cambridge, UK\\
\email{er258@cam.ac.uk}}
\\
Rohan Loveland\at
Department of Engineering Science\\
University of Oxford, Oxford, UK\\
\email{rohan@robots.ox.ac.uk}
}

\maketitle
\input{abstract}

\input{intro}
\input{salientedges}

\input{transform}
\input{hough}
\input{costFunction}
\input{optimizing}
\input{warp}

\input{results}

\input{conclusions}

\bibliographystyle{spmpsci-ed}
\bibliography{papers}
\appendix
\input{jacobian.tex}

\end{document}

%% file: abstract.tex
\begin{abstract}
In this paper we present a simple and robust method for self-correction of camera distortion using single images of scenes which contain straight lines.  Since the most common distortion can be modelled as radial distortion,  we illustrate the method using the Harris radial distortion model, but the method is applicable to any distortion model.  The method is based on transforming the edgels of the distorted image to a 1-D angular Hough space, and optimizing the distortion correction parameters which minimize the entropy of the corresponding normalized histogram.  Properly corrected imagery will have fewer curved lines, and therefore less spread in Hough space.  Since the method does not rely on any image structure beyond the existence of edgels sharing some common orientations and does not use edge fitting, it is applicable to a wide variety of image types. For instance, it can be applied equally well to images of texture with weak but dominant orientations, or images with strong vanishing points.  Finally, the method is performed on both synthetic and real data revealing that it is particularly robust to noise.

\keywords{Radial distortion \and Camera distortion \and Plumb-line constraint}
\end{abstract}

%% file: intro.tex
\section{Introduction}
\subsection{Camera Calibration and Distortion Correction}
Camera calibration addresses the problem of finding the parameters necessary to describe the mapping between 3-D world coordinates and 2-D image coordinates.  This can be divided into the determination of extrinsic and intrinsic parameters.  The extrinsic parameters are necessary to relate an arbitrary 3-D world coordinate system to the internal 3-D camera coordinate system, where the z-axis is typically taken to be along the optical axis.  The intrinsic parameters are independent of the camera position, and typically include a focal length, an offset relating pixel coordinates center to the optical center, a scale factor describing the pixel aspect ratio, and one or more parameters describing non-linear radial and tangential distortions.  Detailed examples of this process can be found in  \cite{heikkila00,lenz88}, among others.

In this work we address the determination of the non-linear distortion parameters, with the assumption that the other intrinsic parameters, usually addressed assuming a pinhole camera model, can be found after the initial non-linear distortion correction.  In order to do this a model must be selected, as well as a method of estimating the model parameters.

\subsection{Summary of Related Work}
\subsubsection{Distortion Correction Models}
A general equation for distortion correction is given by
\begin{equation}
    I_1(\Vec{x}')= I_0(\Vec{x}), \;\;\;  \Vec{x}^{\prime}\equiv \Vec{D(\Vec{x})},
    \label{eq:genDist}
\end{equation}
where $I_1$ is the output (corrected) image, $I_0$ is the input (distorted) image, and $\Vec{x}$ is a pixel location in the input image that is mapped to location $\Vec{x}'$ in the output image by distortion correction function $\Vec{D(\Vec{x})}: \rtwo \Rightarrow \rtwo$.

If the distortion is assumed to be isotropic, and strictly radial, then $\Vec{D(\Vec{x})}$ can be written as
\begin{equation}
    \Vec{D(\Vec{x})} = f(\rho)\hat{\Vec{r}}+\Vec{c},
\end{equation}
where $\Vec{c}$ is the center or radial distortion,
the radius vector $\Vec{r} = [r_1\ r_2]\Trn = \Vec{x} - \Vec{c}$, the radius $\rho = \|\Vec{r}\|$, the  normalized radius vector $\hat{\Vec{r}} = \Vec{r}/\rho$ and $f$ is a scalar function of $\rho$.

Several models for distortion correction have been used previously. In terms of $f$, the models include the polynomial model~\cite{heikkila00,beyer92}, where
\begin{equation}
    f(\rho) = \rho(1 + k_1\rho^2 + k_2 \rho^4 + k_3\rho^6 + ...),
\end{equation}
and the Harris model~\cite{tordoff04}, where
\begin{equation}
    f(\rho) = \frac{\rho}{\sqrt{1+\gamma\rho^2}}.
    \label{eq:Harris}
\end{equation}

More complex models such as~\cite{heikkila97camera} include additional tangential distortion
\begin{equation}
	 \Vec{D(\Vec{x})} = \begin{bmatrix}
	 						p_1(\rho^2 + 2r_1^2) + 2p_2r_1r_2\\
							2p_1r_1r_2 + p_2(\rho^2 + 2r_2^2)
						\end{bmatrix} + f(\rho)\hat{\Vec{r}}+\Vec{c},
\end{equation}
to account for decentering of the optical system (where $p_1$ and $p_2$ are the parameters).
More recently the rational function model, where polynomial and perspective transforms are combined into a rational polynomial~\cite{claus05rational,rosten06accurate} has been proposed.

All of these models can have adequate performance depending on the characteristics of the camera optics.  In the following we limit our discussion and implementation to a single model in order to bound the scope of the paper, though it should be noted that the method we present is extensible to any model. The Harris model is chosen for its reasonable performance, single parameter, and ease of inversion (achieved by simply negating $\gamma$).

\subsubsection{Parameter Estimation Methods}
A variety of methods exist for estimating camera distortion correction model parameters.  Earlier efforts relied on imagery with artificially created structure, either in the form of a test-field, populated with objects having known 3-D world coordinates, or using square calibration grids with lines at constant intervals~\cite{heikkila00,lenz88,beyer92}.  Alternative approaches do not require artificially created structure, but used multiple views of the same scene. The calibration technique makes use of constraints due to known camera motion (for instance rotation)~\cite{stein95}, known scene geometry such as planar scenes~\cite{sawhney99image} or general motion and geometry constrained with the epipolar constraint~\cite{stein96,barreto05fundamental,claus05rational}.

These approaches required access to the camera in order to perform a specific operation, such as acquiring views from multiple positions or views of a particular scene.  This is problematic in instances where access is no longer available to the camera, but only to the resulting imagery.

As a result, a number of methods have been developed which can operate on single views but which make use of common structure in images such as vanishing points~\cite{voss02automatic}, higher-order correlations in the frequency domain~\cite{faird01distortion} and straight lines~\cite{brown71}.

The methods relying on the existence of straight lines, the ``plumb-line''
methods, were pioneered by Brown~\cite{brown71}.  Brown used a test-field with
actual strung plumb-lines, but in general these methods do not require
knowledge of the locations of the lines.  In particular, white plumb-lines were
strung across a black background, with the plumb-bobs immersed in oil for
stability. The photographic plate was exposed twice with the camera rotated
about the optic axis by 90\Deg, giving a nominally square grid of lines. A
number of points along the lines recorded on the photographic plate were
measured using a microscope (a Mann comparator). Calibration was then performed
by minimizing the least-squares error in the image between distorted
straight lines and the measures points on the photographed plumb-lines.

Plumb-line methods are applicable to many image types, because nearly all scenes
containing man-made structures have a large number of straight lines.
Plumb-line methods generally rely on the process of optimizing the distortion
correction parameters to make lines that are curved by radial distortion
straight in the corrected imagery.  The lines can be manually selected, as in
\cite{swaminathan98}, or they can be found automatically, as in
\cite{devernay01,strand05correcting,claus05plumbline} by detecting edgels and
linking them in to line segments.

The objective function for optimization can be formulating by undistorting the line segments
and measuring the straightness by fitting a straight
line~\cite{swaminathan98,devernay01}. Alternatively, the distortion model can be
chosen so that straight lines become specific family of curves, such as
conics~\cite{claus05plumbline} or circles~\cite{strand05correcting}. The
distortion can then be found by fitting these curves to the distorted line
segments.

\section{Algorithm Overview}
We propose a method that is simple and robust to high levels of noise, as shown in the results section.  In our algorithm we calculate all image edgels, and then transform these into a one-dimensional Hough space representation of angle. This creates an orientation histogram of the edgel angles.  In this form, curved lines will be represented at a variety of angles, while straight lines will be found only at one.  Therefore, we optimize the model distortion parameters which minimize the entropy (or spread) of the Hough space angular representation.  The individual steps are:
\begin{enumerate}
\item Find salient image edgels, with normal vectors.
\item Perform a distortion correcting transformation to the edgels.
\item Compute the 1-D angular Hough transform.
\item Compute an objective function defined as the spread (entropy) of the 1-D Hough transform.
\item Optimize the transformation parameters to minimize the entropy/spread based objective function, iterating steps 2--4,
\item Use the optimized transform parameters to map the input image to a corrected output image.
\end{enumerate}

Note that step 1, finding the edgels, is only required once, due to the fact that the edgels, rather than the underlying image, can be transformed.  This, and the other steps of the process, are described in further detail in the following respectively enumerated subsections.

%% file: salientedges.tex
\subsection{Salient edge extraction}

\begin{figure}
\begin{tabular}{ccc}
\includegraphics[width=0.15\textwidth]{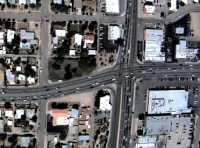}&
\includegraphics[width=0.15\textwidth]{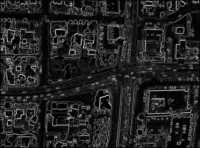}&
\includegraphics[width=0.15\textwidth]{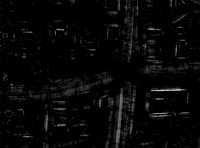}\\
(a) & (b) & (c)
\end{tabular}
\caption{Salient edge detection using tensor voting. (a) The original image, (b)
the gradient magnitude, (c) the edge saliency.\label{fig:tensorvoting}}
\end{figure}

The structure that we are making use of in this paper consists of long, straight
edges, which appear as long, smoothly curved edges in the input image.

Simply performing an edge detection does not result in these features being
dominant, so to enhance the long edges, we use tensor
voting~\cite{lee99grouping} on the dense gradient image. The gradient is
produced by finite differences, and voting is made in proportion to the
gradient magnitude.  We use the standard kernel for smooth, circular curves,
specifically the kernel given by the equations on page 60 of
~\cite{lee99grouping}.

For edges, each pixel
is represented by a $2\times2$ positive semi-definite matrix with eigenvalues
$\lambda_1$ and $\lambda_2$. We wish to find points which are sufficiently
`edgey' so we use the edge saliency function $\phi$, given by:
\begin{equation}
\phi = \max\lambda_1,\lambda_2 - e \min\lambda_1,\lambda_2.
\end{equation}
This detects points where the edge in the main orientation is significantly
brighter than other edges. We have found that $e=2$ produces good results. An
example of these are shown in \Fig{tensorvoting}.
The normal of these edgels is given by the eigenvector corresponding to the
largest eigenvalue.

The final stage is to discard the non-salient edges by thresholding at $\phi=0$.
For computational efficiency of the later stages, the set of edges passing the
threshold is subsampled. Typically, we keep 100,000 edgels. In order to prevent the result being dominated by one region with strong edges, we split the image up in to a regular grid, and perform subsampling independently within each grid cell, so that each cell contains the same number of edgels.

%% file: transform.tex
\subsection{Distortion correction model}

\subsubsection{Anisotropic extension}
As mentioned previously, for the basic radial distortion correction model, we have based our distortion correction model on the Harris model defined in Equation \ref{eq:Harris}.  In order to extend the model's flexibility, we add an anisotropic component, so that the model becomes:
\begin{equation}
	\Vec{D(\Vec{x})} = \hat{\Vec{r}}f(\rho)(1+g(\theta)) + \Vec{c},
\end{equation}
where $f(.)$ and $g(.)$ are the distortion functions.

In particular,
we define the anisotropy function, $g(\theta)$ to be:
\begin{eqnarray}
	g(\theta) &=& a_1 \sin (\theta + \psi_1) + a_2 \sin^2 (\theta + \psi_2) + \ldots\notag\\
	          &=& b_1 \sin \theta + b_2 \cos \theta + (b_3 \sin \theta + b_4\cos\theta)^2 + \ldots,
\end{eqnarray}
which can be rearranged as a Fourier series.  {\markchange This general-purpose
formulation has no discontinuity as the angle wraps around  and conveniently
avoids the use of trigonometric functions, since $\cos \theta = r_1/\rho$,
etc.}

\subsubsection{Edgel transformation}
\def\line{\Vec{l}}
\def\linenormal{\Vec{n}}
\def\tline{\Vec{m}}
\def\tnormal{\Vec{h}}
\def\rtwo{\ensuremath{\mathbb{R}^2}\xspace}
\def\transformation{\Vec{D}\xspace}
\def\trans#1{\ensuremath{\transformation(#1)}\xspace}

\def\B#1{\begin{bmatrix}#1\end{bmatrix}}
\def\pdif#1#2{\frac{\partial#1}{\partial#2}}

In order to evaluate the cost function it is necessary to determine the edgels' orientation, as discussed in the previous section.  If it were necessary to regenerate these by computing a completely new image for each new set of distortion model parameters, however, the optimization process (involving numerous evaluations of different sets of model parameters) would be prohibitively slow. Furthermore, errors due to resampling would be introduced when the image was undistorted at each iteration.

Fortunately, the edgels need to be determined only once, from the input image, due to the fact that the distortion correcting transformations can be applied directly to the edgels, provided that the Jacobians of the transformations are known.  In order to see this, consider an edgel in the input image as part of a parameterised curve $\line(t)$ in \rtwo. The distortion correction transformation, \Vec{D}, can be applied to this space to create a new line:
\begin{equation}
\tline(t) = \trans{\line(t)}.
\end{equation}
The line tangents are then found by differentiating to be:
\begin{equation}
\B{\pdif{m_1}{t}\\\pdif{m_2}{t}} = \Vec{J}
\B{\pdif{l_1}{t}\\\pdif{l_2}{t}},
\end{equation}
where the Jacobian, \Vec{J}, is given by:
\begin{equation}
\Vec{J} = \B{
				\pdif{D_1}{l_l} & \pdif{D_1}{l_2} \\
				\pdif{D_2}{l_l} & \pdif{D_2}{l_2}
			}.
\end{equation}
The edgels include normals \Vec{n} defined at discrete points. {\markchange  The
transformed edgel normals, \Vec{h}, can then be found by rotating the normals
by 90\Deg (making them tangents), transforming the tangents by \Vec{J} and
rotating the transformed tangents back to be normals:} 
\begin{equation}
\tnormal = \B{0 &1\\-1&0} \Vec{J} \B{0 &-1\\1&0} \linenormal
\end{equation}

Note that we do not parameterise the line with a function. The line and its
normal is known (and used) only at a discrete set of points, specifically where
the edgels are detected. This means that $\line(t)$ and $\linenormal(t)$ can be
evaluated at every value of $t$ we require.  Since the edgel detection process
also provides the normals, $\Vec{J}$ is only a function of the distortion
model, and is therefore computed analytically from the definition of $\Vec{D}$. 
The derivation of \Vec{J} for the Harris model is given in Appendix~\ref{sec:jacobian}.

%% file: hough.tex
\subsection{1-D Angular Hough transform}
\begin{figure}
\includegraphics[width=.5\textwidth]{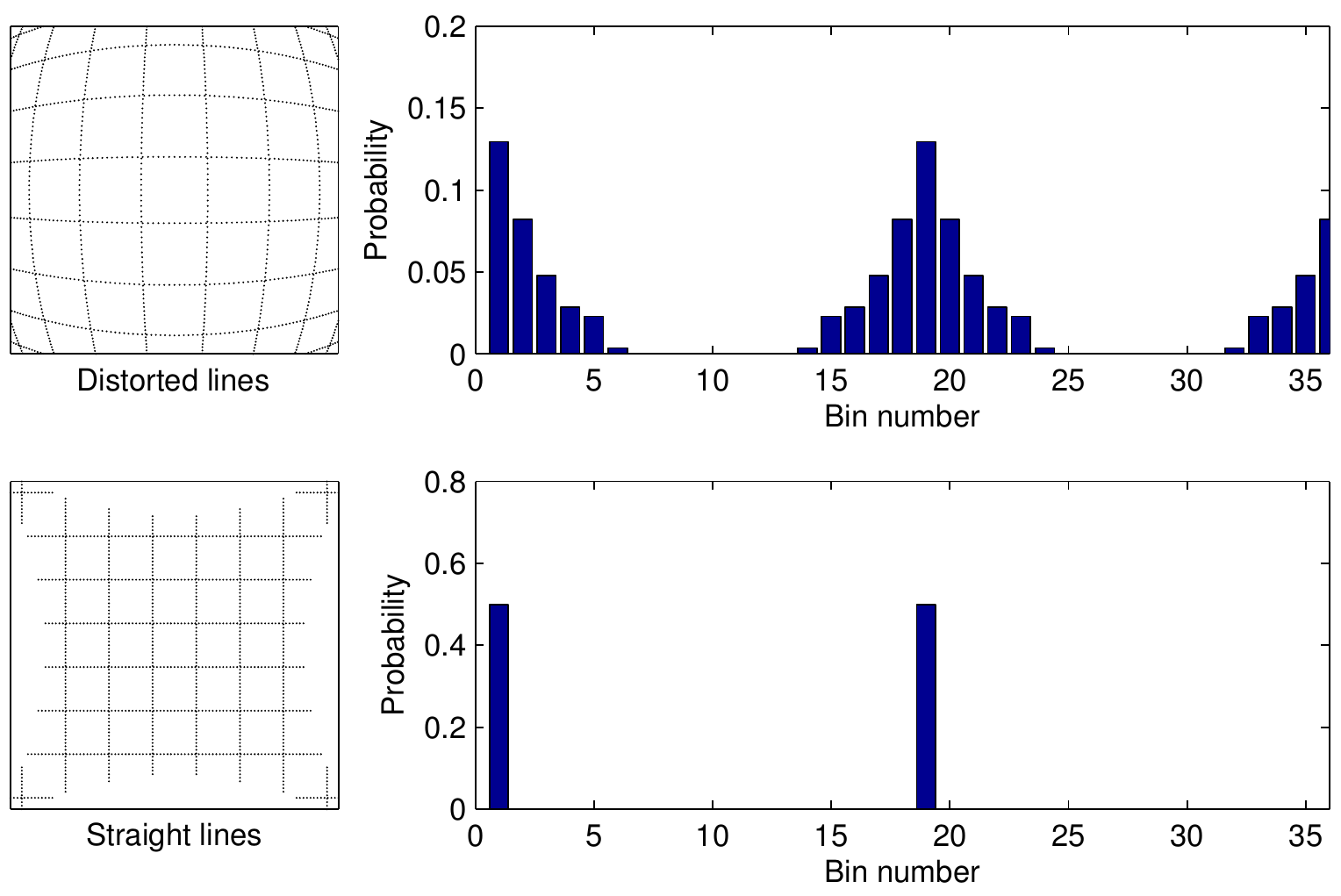}
\caption{Images of curved and straight line images and the corresponding 1-D Hough transforms.\label{fig:Hough}}
\end{figure}

The Hough transform is a technique for finding lines in images~\cite{duda72}.  It is sufficiently well-known so that only a truncated explanation will be given here.  The basis of the technique is the transform of a line to a point in ``Hough'' space.  

By way of example, if a line is defined by $y=mx+b$, then it can be represented by a single point in a 2-D Hough space of $[m]\times[b]$.  An edgel, which contains both a normal vector (and thus the slope of a line) and a discrete point (which a corresponding line would pass through), can also be mapped to a single point in Hough space.  In practice the $mx+b$ formulation is unwieldy, so
\begin{equation}
    \rho=x\cos{\theta}+y\sin{\theta}
\end{equation} is used, where $\rho$ is the shortest distance from the line to the origin, and $\theta$ is defined by the vector normal to the line.

The Hough space can be quantized into discrete bins, and each edgel can be assigned to a particular bin. Given a set of edgels, support for the existence of a particular line is then indicated by the accumulation of a large number of edgels in the corresponding bin.  The edgels from a curve end up in adjacent bins, resulting in a diffuse cluster of non-zero bins, while edgels that are all on the same line end up in exactly the same bin.  This important fact motivates the definition of our objective function, as explained in the next section.

In general we anticipate the presence of a number of parallel lines in the scene, and would like to utilize their reinforcement of each other.  We therefore marginalize the $[\rho]\times[\theta]$ Hough space into a 1-D $\theta$ space by summing over the $\rho$ values for each $\theta$.  An example of some images with curved lines and straight lines, and their corresponding 1-D Hough transforms, are shown in Fig. \ref{fig:Hough}.  Note that the two principal line directions are mapped to two corresponding peaks in Hough space.

%% file: costFunction.tex
\subsection{Entropy-based objective function}

The radial distortion correction method presented here is motivated by the observation that curved lines map to spread out peaks in Hough space, while straight lines map to a single bin.  Therefore, it is desirable to have an objective function that measures this spread.  In information theory this quality is represented by entropy~\cite{shannon48mathematical}.  We have therefore normalized the 1-D Hough representation, and treat it as a probability distribution.  The objective function is then:
\begin{equation}
\displaystyle
     C(\Vec{H})\equiv-\sum_{b=1}^{B} p(H_b)log_2(p(H_b)),
\end{equation} 
where \Vec{H} is the discretized 1-D angular Hough transform of the edgels of an input image for a given set of model parameters, $p(H_b)$ is the value of a given normalized Hough bin, and $B$ is the number of bins. Minimizing $C$ therefore minimizes the spread.

%% file: optimizing.tex
\subsection{Optimization}

The cost function has many local minima, so to optimize it effectively a
reasonable strategy is needed. As a broad overview, we have found that the MCDH
(Monte-Carlo downhill) strategy works best. Essentially, starting parameters
are selected at random, and the downhill simplex
algorithm~\cite{downhill-simplex} (specificially the variation given in
\cite{nr_in_C_opt}) is used to optimize the cost from the selected starting
points. The best result is then selected.  The downhill simplex algorithm is
iterated until the residual drops below a threshold ($10^{-15}$), or 1000
iterations occur, whichever is sooner. The residual is computed as:
\[
	r = \frac{\|\Vec{s}_\text{best} -
	\Vec{s}_\text{worst}\|_2}{\|\Vec{s}_\text{best}\|_2},
\]
where  $\Vec{s}_\text{best}$ is the location of the best (smallest) point on the
simplex, and $\Vec{s}_\text{worst}$ is the location of the worst (largest point)
on the simplex.

In order to implement this method effectively, several techniques are needed.
Parameters computed in the Hough transform will not lie exactly at the bin
centres. If the parameter is simply placed in the closest bin, then the
quantization caused by this causes flat areas to exist in the cost function.
The flat areas can easily confound the optimizer, so instead, the neighbouring
bins are incremented using linear interpolation.

The optimizer is most effective when the parameters being optimized have similar
orders of magnitude, since otherwise roundoff error will cause errors in the
computation of new simplexes. In this case, a simple scale based on
\emph{a priori} knowledge of approximate parameter values suffices.

For radial distortion, \centre is roughly in the middle of an image and is of
the order $10^2$ to $10^4$ pixels. Given Equation~\ref{eq:Harris}, it is
reasonable to expect $\gamma$ to be approximately in the range
$\pm\rho_\text{max}^{-2}$, where $\rho_\text{max}\approx |\tilde\centre|$ and
$\tilde\centre$ is the centre of the image.
Therefore, in
the optimization, we solve for $\beta$, where $\beta =
100\gamma\rho_\text{max}^{2}$, which brings $\beta$ in to the same range as
$\centre$.

Similarly, we have found that the values of the anisotropic coefficients,
$b_1, \ldots$ are of order $10^{-3}$, so we instead solve for $d_i$, where $d_i =
10^5b_i$. The resulting optimization is performed over the vector:
\begin{equation}
	[ c_1, c_2, \beta, d_1, d_2, d_3, d_4, d_5, d_6 ].
\end{equation}

The random selection of the starting parameters must be based around
some knowledge of the values. The centre of radial distortion is usually near to
the centre of the image, so we draw the initial centres from a normal distribution
centred at $\tilde\centre$ with a standard deviation of
$\frac{|\tilde\centre|}{20}$. Given the approximate values of $\gamma$, and
therefore $\beta$, we draw initial values of $\beta$ from  $\mathcal{N}(0,
100)$. The initial anisotropy parameters are set to zero.
For our application and these parameters, we have found that 120 MCDH
iterations is sufficient.

%% file: warp.tex
\subsection{Generating the output image}

In order to generate the output (corrected) image the following procedure is needed:
\begin{enumerate}
\item For each point, \Vec{x'}, in a grid defined on the \emph{output} image $I_1$, find the
corresponding point in the distorted input image $I_0$ given by $\transformation^{-1}(\Vec{x'})$.
\item Copy the pixel back, performing the assignment:
\[
	I_1(\Vec{x}') := I_0(\transformation^{-1}(\Vec{x}')).
\]
\end{enumerate} 

Since images are discrete, interpolation will be required to find $I_1(\transformation^{-1}(\Vec{x}'))$.  Bicubic interpolation has been used to generate the results shown in the following section.

%% file: results.tex
\section{Results}

\subsection{Synthetic tests}

\begin{figure}
\includegraphics[width=0.5\textwidth]{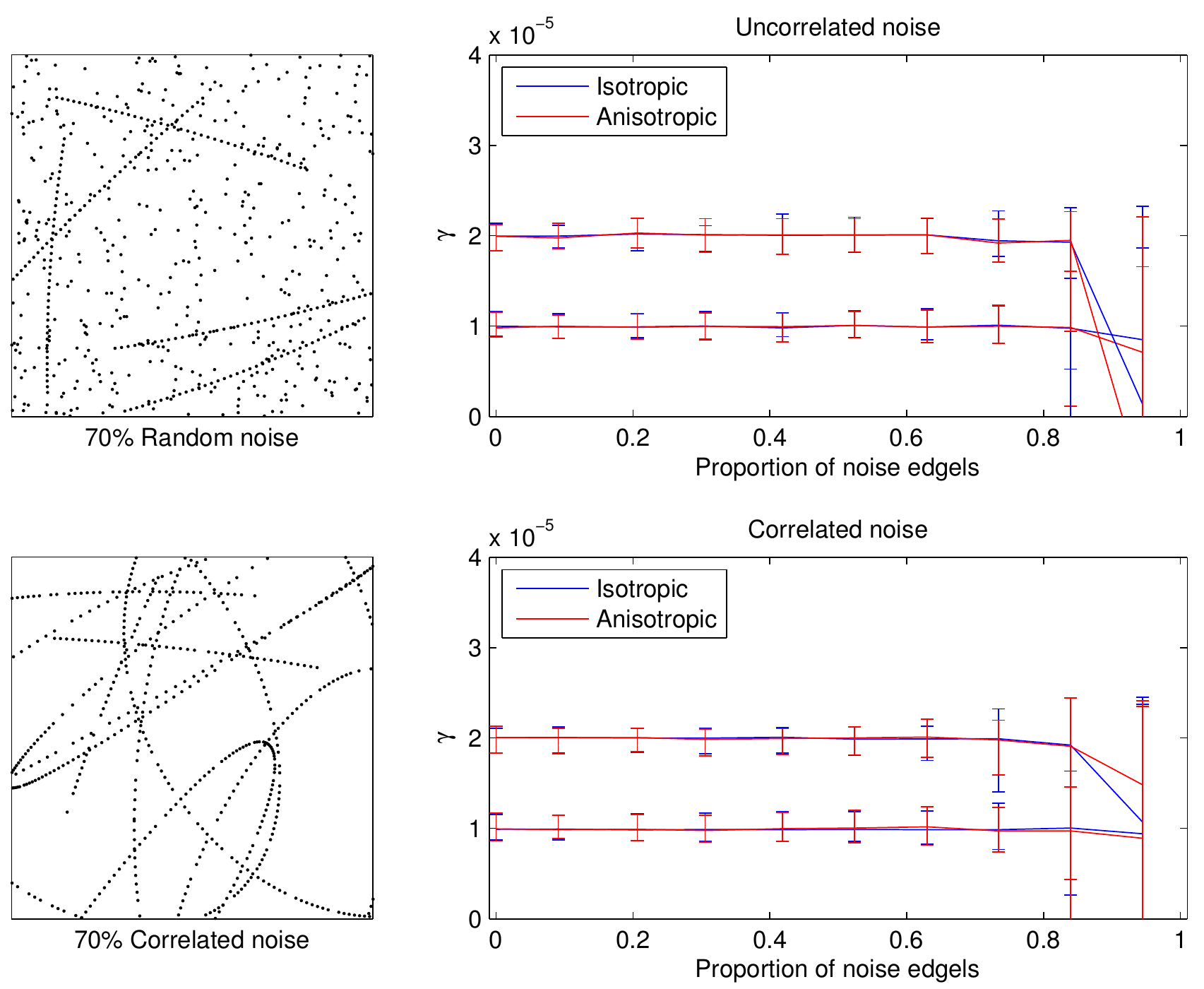}
\caption{(Left) Illustration of the simulated data (shown with 100 points per
line).  (Right) Performance of the algorithm on the simulated data, with
$\gamma=10^{-5}$ and $\gamma=2\times10^{-5}$.  The median value of gamma is
shown with error bars at the 10$^\text{th}$ and 90$^\text{th}$ percentile.
\label{fig:noise}}
\end{figure}

In order to compute the sensitivity to various kinds of noise, we create a
number of synthetic images, using known radial distortion levels, and compute
the resulting radial distortion correction parameters using our technique.  We
investigate the two correction models, namely the strictly radial correction of the
unmodified Harris model, and the anisotropic extension discussed previously. The
images are designed to test the performance of the algorithm in the presence of
measurement error in the plumb-lines, and both uncorrelated and correlated
non-plumb-line data.

The $250\times250$ pixel test images consist of five good quality lines (no
point can be within 60 pixels of the centre of the image), consisting of 10
points on each line, drawn with barrel distortion and with random orientation
errors.  Then (in the case of uncorrelated noise), a number of randomly placed,
randomly oriented points are added. In the case of correlated noise, a number
of randomly placed, sized and oriented ellipses (with the same approximate
point spacing as the lines) are added. Our algorithm is then applied to the
test image to estimate the parameters.

For every selected noise proportion for the simulation, 200 test images are
created for each of two different values of $\gamma$. The results of these, and
an illustration of the simulated data is shown in \Fig{noise}. As can be seen,
the technique is very robust to noise, producing good results with up to 70\%
contamination and interestingly it does not perform much worse in the case of
correlated noise.  Furthermore, although the anisotropic extension (in this
case) introduces six extra parameters to the optimization, this has no
significant effect on the stability of the technique in high noise situations.

\subsection{Example images}

\begin{figure*}
\begin{tabular}{cccc}
\includegraphics[width=0.45\textwidth]{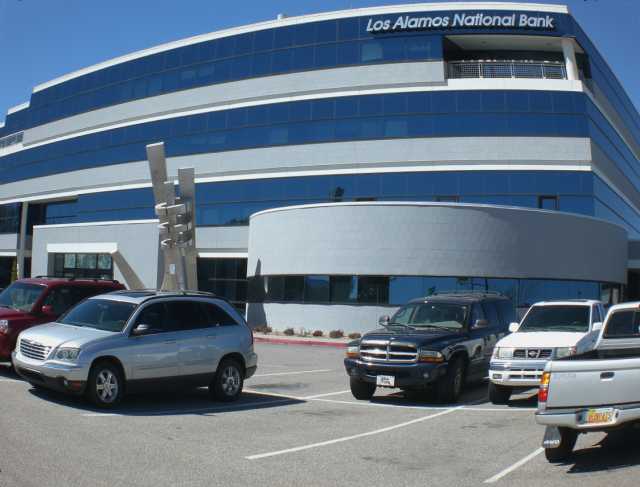}&
\includegraphics[width=0.45\textwidth]{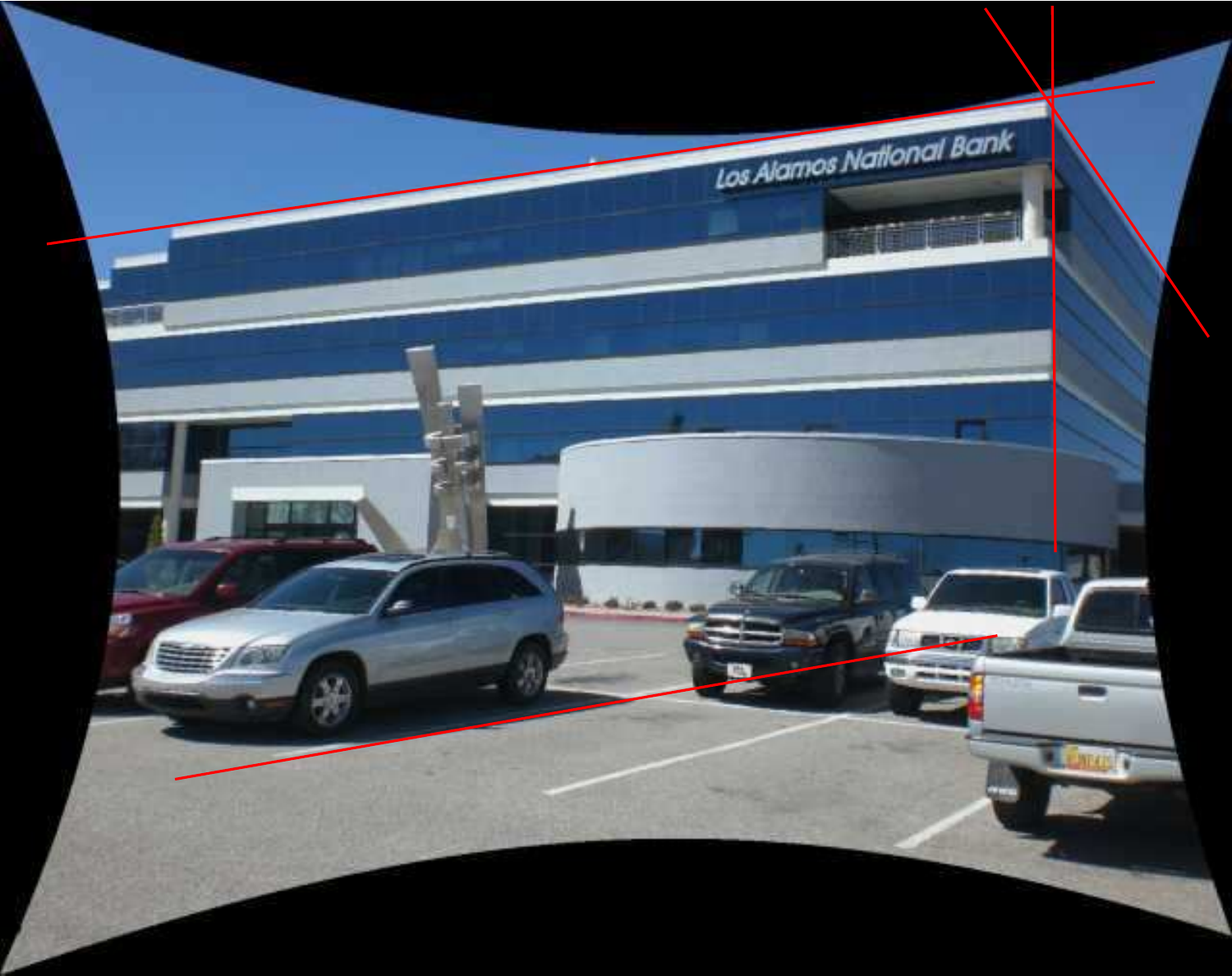}\\
\includegraphics[width=0.45\textwidth]{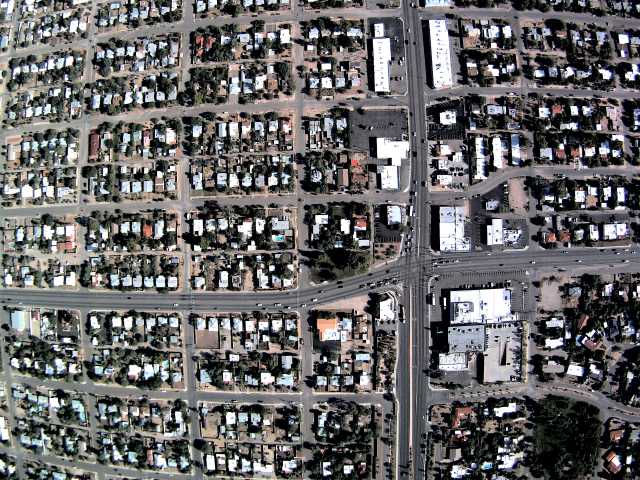}&
\includegraphics[width=0.45\textwidth]{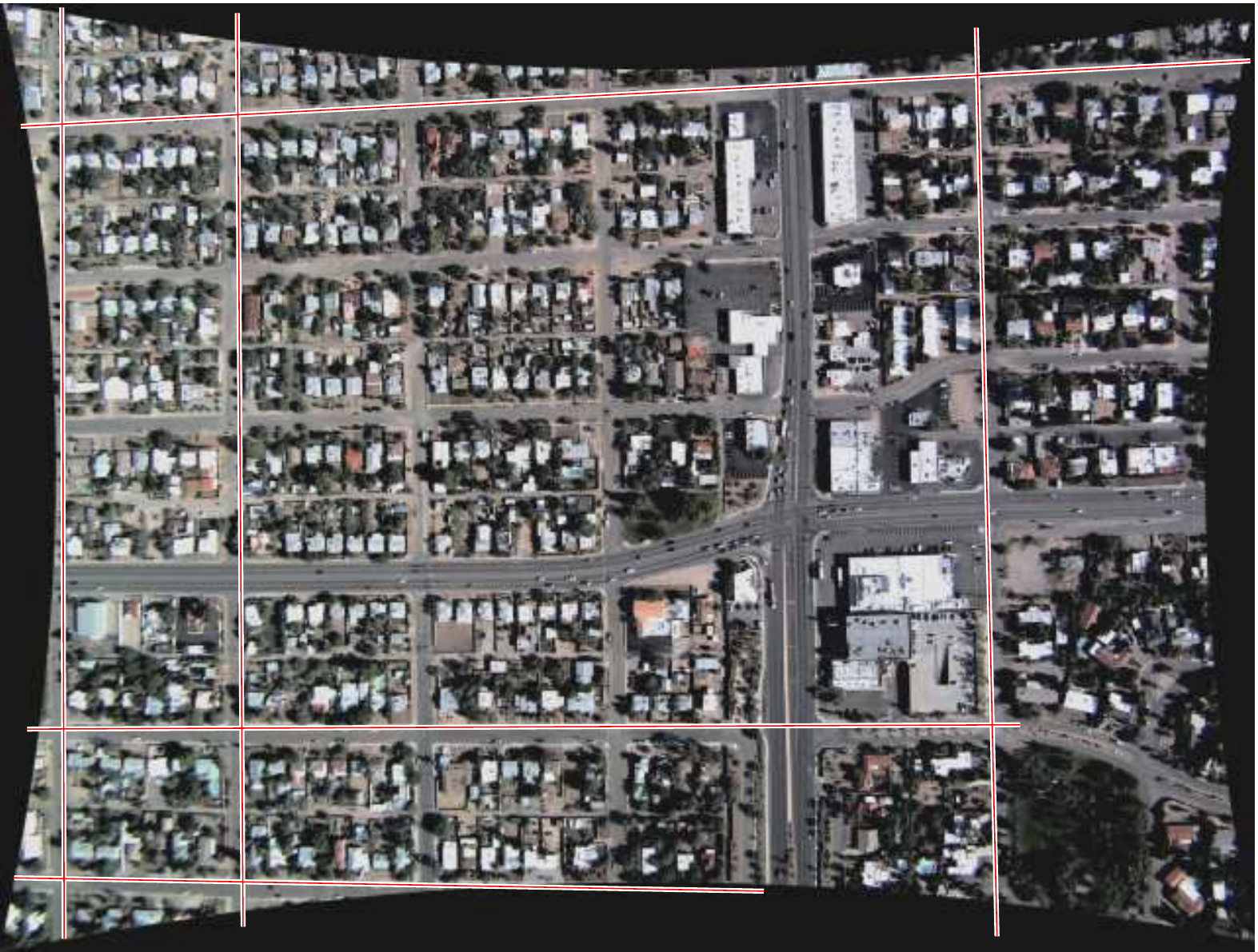}
\end{tabular}
\caption{(Left) Original, distorted images. (Right) Images undistorted using
our technique.  Some lines which are straight in the world have been annotated
with straight lines in the undistorted image.\label{fig:real}
The images are of a building, showing strong vanishing points and
an aerial image of a city.}
\end{figure*}

The results on two real images are shown in \Fig{real}. In the corrected images,
real-world straight lines have been annotated with lines to show that there is
no significant amount of curvature. The upper image shows the technique
operating on an image where the lines belong to vanishing points (note that some
distortion was artificially added to this image in addition to the camera
distortion present). Note the presence of strong curved edges along the
boundaries of the cars and parts of the building.

The lower image shows the results on an aerial image of a
city. Although there are two principal directions in this image, the contrast on
the principal edges is low compared to many of the other features.
Additionally, the anisotropy extension was required to correct the very strong
distortion in the lower left corner.

\subsection{Comparison to other techniques}

To gauge the accuracy of our method, we have compared it to a
technique based on structured scenes. In particular we have used the technique
of~\cite{libcvd-calibration-calibpaper}, which uses multiple images of a planar
grid of squares to determine the intrinsic and extrinsic camera calibration
parameters. This system is able to calibrate cubic, quintic, Harris and Harris
with unit aspect ratio camera models. We use the last listed model, since this
matches the camera model in the paper.  In particular, we use this program to
optimize the model:
\begin{equation}
\def\BrkMat#1{\begin{bmatrix}#1\end{bmatrix}}
\BrkMat{x_i\\y_i} = \BrkMat{u\\v} +  \BrkMat{f& 0\\0&f}
                    \BrkMat{x_c\\y_c} 
					\frac{1}{\sqrt{1 + \alpha (x_c^2 + y_c^2)}},
\end{equation}
where $(u,v)$ is the optic axis, $(x_i, y_i)$ is a coordinate in the image, and
$(x_c, y_c)$ is a coordinate in ideal, normalized camera's image plane. Since
the technique is a structured scene technique, it also estimates the focal
length of the camera, $f$. To translate from one model to the other, set $\gamma =
\frac{\alpha}{f^2}$.  Note that this assumes that the centre of radial
distortion is at the optic axis of the camera, which is a reasonable
assumption.

In order to measure the quality of the camera calibration, we localise a known
3D model in an image, and measure the root mean square (RMS) error between the
control points on the model and the measured points in the image. The model
consists of 13 by 9 black and white squares in a chequerboard pattern on a flat
plane. The model is warped using a homography (the parameters of which must be
determined), then by the calibrated radial distortion function and rendered in
to the image. We then search normal to the rendered edges, looking for edges in
the image. 
We define the error to be the distance offset in pixels between the rendered
model edge and the measured edge position in the image
The parameters of the homography are then adjusted
using the downhill-simplex algorithm to minimize the RMS error. In particular,
we search 15 pixels in either direction, and take the point with the highest
gradient, measured using the kernel $[-1\ -2\ -1\ 0\ 1\ 2\ 1]$. If the highest
gradient does not exceed some threshold, or is not a local maxima, then the
edge search is discarded and does not contribute to the RMS error. Quadratic
interpolation is then used to find the edge position to subpixel accuracy.  The
search distance is sufficiently small that data association is trivial, and
robust fitting is not required.  A more detailed treatment of very closely
related systems for aligning models is given in~\cite{rapid,Drmtrack}.

\begin{figure*}
Test images:

\begin{tabular}{ccc}
\includegraphics[width=0.3\textwidth]{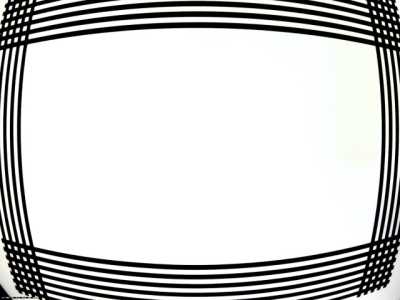} &
\includegraphics[width=0.3\textwidth]{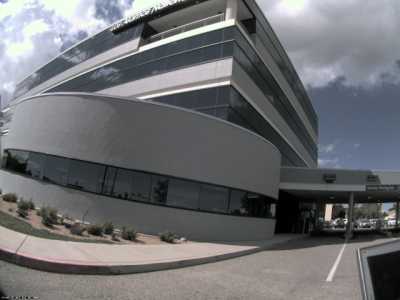} &
\includegraphics[width=0.3\textwidth]{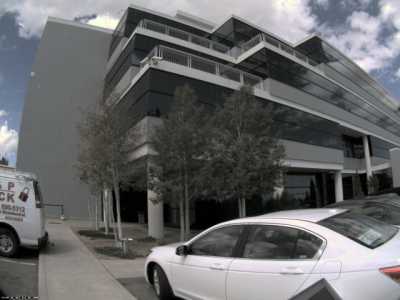}\\
1 & 2 & 3 \\
\includegraphics[width=0.3\textwidth]{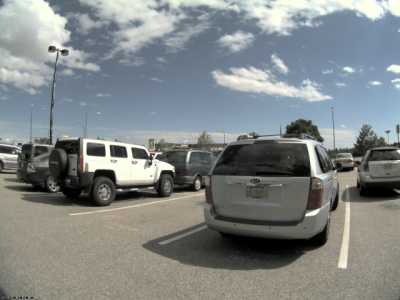} &
\includegraphics[width=0.3\textwidth]{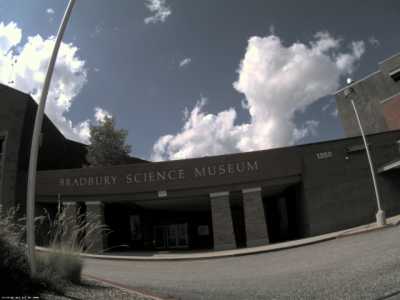} &
\includegraphics[width=0.3\textwidth]{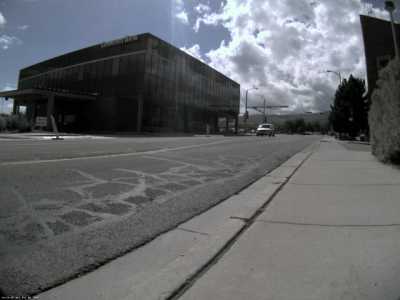} \\
4 & 5 & 6
\end{tabular}

Calibration results:

\begin{tabular}{l|rrr|r@{}l}
Calibration               &       $c_1$      &     $c_2$        & $\gamma\times10^{7}$  & RMS error\\\hline
Uncalibrated              &   1024           &     768.0        &     0.000             &    $>40\ddagger$ \\\hline
libCVD camera calibration &   1030           &     687.0        &     3.780             &    1.67 \\\hline
Test image 1              &   1075           &     708.6        &     3.700             &    2.38 \\
Test image 2              &   1325           &     769.0        &     3.408             &    12.2&$\dagger$ \\
Test image 3              &   1149           &     627.3        &     4.317             &    4.75 \\
Test image 4              &   1012           &     739.1        &     2.814             &    6.08 \\
Test image 5              &   1084           &     611.4        &     3.544             &    3.60 \\
Test image 6              &   1226           &     968.0        &     3.405             &    12.8&$\dagger$ \\
Test images 6 and 2       &   1047           &     725.6        &     4.075             &    2.21 \\
\end{tabular}
\caption{Calibration results for an alternative method and some test images,
including a calibration grid and some images of urban scenes. All images
are $2048\times1536$ pixels.
$\ddagger$ In the
uncalibrated case, the errors are sufficiently that the search radius had to be
increased to 100 pixels. This results in the RMS error being an underestimate,
since for large errors, the system sometimes finds a closer edge (so the
measured error is too small), or fails to find the edge at all (so large errors
are not included).
$\dagger$ In some cases, the errors caused a large number of the correspondences
to be missed with a search radius of 15 pixels, so it was increased to 36
pixels. For the last row of the table, the edgels from the two images were
aggregated and treated as a single image.\label{fig:comparison} In our model,
$\Vec{c} = [c_1\ c_2]$ and for the libCVD model, $u = c_1$ and $v = c_2$.
}
\end{figure*}

The results of this test are shown in \Fig{comparison}. As expected, the
technique which uses many (in this case approximately 100) images of a
structured scene performs best. The performance of our method varies somewhat
with the contents of the scene, but performs well on certain images. One
interesting property of our technique is well illustrated. The property is that
edgels from multiple, unrelated images can be aggregated to provide more
straight edges. In the case shown, the resulting calibration is superior to the
calibration generated from a single image of a carefully constructed grid.

%% file: conclusions.tex
\section{Discussion and conclusions}

In this paper, we have presented a new, simple and robust method for
determining the radial distortion of an image using the plumb-line constraint.
The technique works by first extracting salient edgels and then minimizing
the spread of a 1D angular Hough transform of these edgels. The technique is
simple and because no edge fitting is performed, the technique is very robust to
the presence of noise. Furthermore, the technique is more generally applicable
than other plumb-line techniques in that the lines used do not need to be
continuous. The technique works on textures with principal directions, as
illustrated by the aerial image of a city, where the salient edge detection
results in a large number of relatively small edge fragments.

The proposed algorithm has a number of parameters: the parameters of the tensor
voting kernel, the number of bins and the parameters of the optimization. In
practice, the selection of these parameters are not critical, and indeed the
same set of parameters was used for the simulated data, the example images
and the test
images shown. 

Our method is flexible in that it does not impose constraints beyond
the presence of one or more straight edges: it is not a requirement that the
edges share vanishing points, or structure of any particular kind. It is not
even a requirement that the edgels belong to a related set of images. The
technique can be equally applied to edgels from multiple images of unrelated
scenes taken with the same camera parameters. Finally, our method is widely
applicable because it is, in terms of RMS error, able to produce a calibration
to within three percentage points of a technique requiring access to the camera
and structured scenes.

%% file: jacobian.tex
\section{Derivation of \Vec{J} for the Harris model}
\label{sec:jacobian}

\def\Mat#1{\ensuremath{\mathbf{#1}}\xspace}
\def\BM#1{\ensuremath{\begin{bmatrix}#1\end{bmatrix}}}
\def\D{\Mat{D}}
\def\X{\Vec{x}}
\def\R{\ensuremath{\hat{\Vec{r}}}\xspace}
\def\Dx{\ensuremath{\D(\Vec{x})}}
\def\Pd#1#2{\ensuremath{\frac{\partial#1}{\partial#2}}}

Taking \R to be a column vector (i.e.\ $\R = \BM{\hat{r}_1\\\hat{r}_2}$):
\begin{align}
\Mat{J} = \Pd{\D}{\X} =& \Pd{\R}{\X} f(\rho)(1+g(\theta))
					+ 
			  \R(1 + g(\theta))\Pd{f(\rho)}{\rho}\Pd{\rho}{\X}
					\notag\\
			&\hspace{7.63em}+\R f(\rho)\Pd{g(\theta(\X))}{\X}\label{eqn:main}
\intertext{Where:}
\Pd{\R}{\X} &= 
	 \BM{\Pd{\hat{r}_1}{x_1} & \Pd{\hat{r}_1}{x_2} \\
		 \Pd{\hat{r}_2}{x_1} & \Pd{\hat{r}_2}{x_2}} = {\markchange
		 \frac{1}{\rho}\BM{\hat{r}_2^2 & -\hat{r}_1\hat{r}_2 \\
                             -\hat{r}_1\hat{r}_2 & \hat{r}_1^2 },\label{eqn:rhatdiff}}\\
\Pd{f(\rho)}{\rho} &= (1 + \gamma\rho^2)^{-\frac{3}{2}},\label{eqn:fdiff}\\
\Pd{\rho}{\X} &= \BM{\Pd{\rho}{x_1}&\Pd{\rho}{x_2}} = \BM{\hat{r}_1 & \hat{r}_2}\label{eqn:rhodiff},
\intertext{and:}
\Pd{g}{\X} &= \BM{\Pd{g}{x_1}&\Pd{\rho}{x_2}} 
\end{align}
The function $g(\theta)$ is a function of $\sin\theta$ and $\cos\theta$, where:
\begin{align}
	\cos \theta &= r_1/\rho = \hat{r}_1, \text{and}\\
	\sin \theta &= r_2/\rho = \hat{r}_2, 
\end{align}
we can rewrite $g(\theta(x))$ as:
\begin{equation}
g(\R) = b_1\hat{r}_2 + b_2\hat{r}_1 + (b_3\hat{r}_2 + b_4\hat{r}_1)^2 + (b_5\hat{r}_2 + b_6\hat{r}_1)^3,
\end{equation}
so,
\begin{equation}
	\Pd{g}{\X} = \Pd{g(\R)}{\R}\Pd{\R}{\X} = \BM{\Pd{g(\R)}{\hat{r}_1} & \Pd{g(\R)}{\hat{r}_2}} \Pd{\R}{\X}\label{eqn:gdiff}
\end{equation}
where:
\begin{align}
\Pd{g(\R)}{\hat{r}_1} &= b_1 + 2b_3(b_3\hat{r}_1 + b_4\hat{r}_2) + 3b_5(b_5\hat{r}_1 + b_6\hat{r}_2)^2\label{eqn:gdiff1}\\
\Pd{g(\R)}{\hat{r}_2} &= b_2 + 2b_4(b_3\hat{r}_1 + b_4\hat{r}_2) + 3b_6(b_5\hat{r}_1 + b_6\hat{r}_2)^2\label{eqn:gdiff2}.
\end{align}
Substituting Equations \ref{eqn:rhatdiff}, \ref{eqn:fdiff}, \ref{eqn:rhodiff}, \ref{eqn:gdiff}, \ref{eqn:gdiff1} and \ref{eqn:gdiff2} in 
to Equation \ref{eqn:main} gives the analytic solution for the \Mat{J}.